# Opioid Named Entity Recognition (ONER-2025) Using Pre-trained Transformers on Reddit Data

Muhammad Ahmad, Member IEEE, Rita Orji, Member IEEE, Fida Ullah, Ildar Batyrshin, Senior Member IEE, and Grigori Sidorov

***Abstract*—The opioid overdose epidemic remains a critical public health concern, particularly in the United States, where it contributes to substantial mortality and social costs. Social media platforms contain rich, unstructured user-generated data reflecting public perceptions and self-reported experiences with opioid use. However, extracting meaningful insights from such data demands advanced Natural Language Processing (NLP) techniques, and to the best of our knowledge, no publicly available dataset existed that focused on domain-specific named entities relevant to the opioid overdose crisis—particularly those capturing how individuals consume opioids through various routes of administration. To fill this gap, this study makes the following four key contributions: 1) we created ONER-2025, a novel manually annotated dataset sourced from Reddit, comprising 331,285 tokens labeled with eight domain-specific entities related to route of administration. 2) We built the annotation guidelines to ensure consistency, and we documented the challenges faced during annotation, such as the presence of slang, ambiguous terms, and fragmented sentences common in informal social media discourse. 3) We conducted 11 different experiments using 5-fold cross-validation, employing a mix of machine learning, deep learning, and transformer-based language models to find the best-fit model. Based on the results, transformer-based models, particularly BERT-base-NER and RoBERTa-base (F1-score = 0.97), significantly outperformed traditional baselines, reflecting a substantial 10.23% improvement over the baseline (RF = 0.88). 4) Finally, we applied a t-test to determine whether the performance improvement was statistically significant or occurred by chance. The t-test results confirm that the improvement is statistically significant.**

## I. Introduction

THE opioid crisis has become a significant global public health issue in many countries, particularly in the United States from 2010 to 2015 [1], where it has claimed hundreds of thousands of lives due to opioid overdoses. The opioid pandemic not only affects individuals and families but also places a tremendous burden on healthcare systems and economies. In 2013, the societal costs associated with prescription opioid overdoses, abuse, and dependence in the United States were estimated to be $78.5 billion [2] and Centers for Disease Control and Prevention (CDC) reports an average of 128 lives daily in the US [13]. Of these, 32% are attributed to overdoses involving legally obtained prescription opioids [14].

In recent years, the rapid growth of social media platforms has led to an unprecedented volume of user-generated, real-time content, offering a valuable yet often underutilized resource for monitoring public discourse. These platforms capture rich, unstructured data reflecting the public's perceptions, personal experiences, news, and discussions related to opioids [3-10], including drug use [21-23], addiction, and public health concerns. Such data can provide immediate insights into the ongoing opioid crisis by revealing patterns of opioid abuse, routes of administration, symptoms, dosages, locations, and individuals involved. Leveraging this wealth of information requires advanced analytical approaches. The increasing availability of textual data, combined with advancements in deep learning (DL) models, has facilitated the development of methods capable of extracting and analyzing healthcare data [11, 12] to address various aspects of the opioid epidemic. DL offers powerful tools for managing complex interactions in large datasets, uncovering hidden patterns, and generating actionable predictions in clinical and public health settings, often outperforming traditional statistical techniques [15-20]. In this context, Natural Language Processing (NLP), and specifically Named Entity Recognition (NER), plays a key role in identifying and categorizing opioid-related entities from social media data to support accurate, timely, and actionable interventions.

The rise of illegal opioid sales and usage on social media platforms presents serious challenges across multiple domains crucial to public health [33]. Drug dealers openly advertise and distribute opioids like fentanyl, heroin, and oxycodone, often using slang terms, code words, and ambiguous language to avoid detection and censorship, which increases accessibility and misuse among vulnerable populations [34]. At the same time, opioid users engage in high-risk behaviors, employing

*(Corresponding author: Grigori Sidorov)*

Muhammad Ahmad, Fida Ullah, Ildar Batyrhin and Grigori Sidorov are with the Centro de Investigación en Computación, Instituto Politécnico Nacional (CIC-PN), Mexico City 07738, Mexico (e-mail: mahmad2024@cic.ipn.mx, fidaullahmohmand@gmail.com, batyr1@cic.ipn.mx, sidorov@cic.ipn.mx).

Rita Orji is with the Faculty of Computer Science, Dalhousie University, 6050 University Avenue, P.O. Box 15000, Halifax, NS B3H 4R2, Canada (e-mail: rita.orji@dal.ca).

various routes of administration such as injection, snorting, and smoking, which significantly elevate the risk of overdose and long-term health complications. Monitoring these complex interactions requires identifying not only drug names but also user behaviors, symptoms of overdose, dosages, and treatment responses. NER is a vital tool in this context, enabling automated extraction and classification of such diverse health-related entities from large-scale textual data, including social media posts, clinical notes, and news reports. This detailed information supports healthcare providers, researchers, and policymakers in tracking trends, understanding risk factors, and developing timely interventions to reduce opioid-related harm. Without precise tools like NER, critical signals in vast unstructured data could be overlooked, limiting effective responses to the opioid crisis. Therefore, incorporating advanced NLP methods into healthcare analytics is essential for improving opioid misuse surveillance, prevention, and treatment strategies.

Several studies have already explored NLP tasks and Named Entity Recognition (NER) in both English and low-resource languages using various machine learning and transformer-based approaches [32]. Prior work such as NER-RoBERTa has demonstrated the feasibility of adapting models for low-resource contexts. NER generally employs three techniques to extract relevant entities from textual data: rule-based, machine learning-based, and hybrid approaches. Each technique has its strengths and limitations. Rule-based techniques focus on predefined linguistic patterns and domain-specific lexicons to identify entities, while machine learning-based techniques learn patterns from labeled training datasets to predict entities, offering better generalization to hidden patterns but requiring high-quality datasets. Hybrid approaches combine rule-based and machine learning methods, leveraging the strengths of each to balance flexibility and accuracy while addressing domain-specific challenges. However, in our research, we are specifically interested in applying NER to a domain-specific setting—opioid-related discussions on Reddit—and particularly exploring machine learning methods to address this task using the ONER-2025 dataset. Unlike general-purpose NER tasks, this domain requires identifying complex and context-specific entities such as drug names, dosages, symptoms, and administration methods. Therefore, our focus is not on multilingual or cross-lingual generalization, but on enhancing entity extraction performance within this critical public health domain.

Therefore, the application of NER in the context of opioid-related discussions is a relatively new approach and yet underexplored area of research. To contribute to this field our study aims to address this gap by leveraging ONER-2025 dataset to identify and analyze opioid-related content on Reddit, especially focusing on people using specific routes of administration. The reason for choosing Reddit is that it is a popular social media platform due to its unique structure that encourages open discussions and community-driven content on subreddits, and it is ranked as the fifth most visited website in the United States. On Reddit, text posts can be up to 40,000 characters long, allowing for detailed discussions within the platform's flexible and user-friendly structure.

To tackle this issue, we manually annotated a dataset using a hybrid approach that combined BioBERT-based with human verification and correction, resulting in high-quality labels across 8 predefined categories: Drug Name, Dosage, Route of Administration, Symptoms, Temporal, Location, Event, and Others. In addition, we employed and evaluated 5-fold cross-validation with machine learning models, including Support Vector Machine (SVM), Random Forest (RF), and Logistic Regression (LR), using token-based feature extraction. Additionally, we employed advanced deep learning models, such as BiLSTM and CNN, with pre-trained word embedding's using FastText and GloVe. Furthermore, we incorporated four language models such as bert-base-NER, roberta-base, biobert-base-cased, and Bio-ClinicalBERT with advanced contextual embedding's to achieve a more accurate identification of opioid-related entities in the our ONER-2025 dataset. Our goal is to develop and advanced Artificial intelligence model which automatically extract clinical entities and analyze their frequency and geographical distribution over time. By doing this research, we aim to develop a robust ONER-2025-Dataset that can support public health surveillance efforts and aid policymakers in detecting early signs of opioid misuse, enabling timely and targeted interventions.

This study makes the following key contributions:

- Manually Annotated Dataset: We created ONER-2025, a unique, manually annotated dataset sourced from Reddit, comprising 331,285 tokens and annotated with eight key categories of named entities specifically designed for Opioid Named Entity Recognition. To the best of our knowledge, ONER-2025 is the first dataset of its kind to address the opioid crisis through NER, featuring a novel set of entity types carefully curated for this domain and not covered by any existing NER corpora;
- Detailed Annotation Process: We described our annotation process and guidelines in detail and discuss the challenges of labeling during dataset construction;
- Contextual Understanding of Social Media Discourse: The tool's use of advanced NLP models allows for a deeper understanding of the context behind opioid-related discussions, addressing challenges like slang, ambiguity, and fragmented sentences for more accurate detection;
- Improved Public Health Response: We proposed a real-time monitoring system designed to process streaming data from social media, healthcare records, and emergency services to identify opioid overdose events, with a focus on the route of administration used by the opioid user. The system leverages 5-fold cross validation in machine learning models using token-based feature extraction, deep learning models with pre-trained embedding's (FastText, GloVe), and transformer-based models with advanced contextual embedding's to address dynamic meanings and contextual relationships within the textual data. This tool enables faster, more informed public health interventions by providing actionable insights for healthcare professionals and policymakers;

- Based on experiments, our proposed model (RoBERTa-base), achieved an F1-score and accuracy of 97%, showing a substantial 10.23% improvement over the baseline (RF =F1-score: 0.88). To confirm that this performance gain was not due to chance, we conducted a paired t-test on the F1-scores, which produced a p-value below 0.05, demonstrating that the improvement is statistically significant.

### A. Challenges in Opioid NER

Clinical NER faces many challenges due to unstructured, informal, and often emotionally charged social media data. Slang usage is a major difficulty, as user often mention drug in non-standard forms which makes it difficult to automatically detect the NER for examples: a person posted a post on Reddit, “Looking for blues tonight, who’s got the hookup?.” In this post, "blues" refers to blue Oxycodone pills. Without prior knowledge of drug-related slang, it is very difficult to detect such terms in NER, which could lead to misinterpretation of the results. Ambiguity further complicates ONER-2025 detection, as many words have several meanings. A person posted, “Need some oxy ASAP”. In this post user want to request for an Oxycodone or an oxygen supply in a medical emergency. Similarly, another user post is “I just got my script refilled for oxycodone, but I'm worried about overdosing again. I usually take it orally, but last time I snorted some and it hit me too hard”. In this post the word is used script could mean a prescription but might also refer to a movie or software script depending on the context. Another challenge on social media data in ONER-2025 is Fragmented sentences, where different users communicate in different ways. For example a post “Got some percs... took too many earlier today, feeling a little off”. DM if needed, I might need help, which is missing clear subjects and objects, relying on the reader’s understanding of opioid-related discourse. The lack of full sentence structure forces ONER systems to mix meaning from minimal context, often leading to errors. Additionally, another common challenge in opioid-related discussions, is emotionally charged language, particularly among individuals struggling with addiction or chronic pain. A person wrote a post on Reddit: "I can’t deal with this pain anymore. Just need to feel numb for a while." While the user did not mention opioids directly, the statement strongly suggests drug use and a desire for relief through substances. These combination of challenges demands advanced Opioid named entity recognition approaches that go beyond simple keyword matching, requiring contextual understanding and adaptability to evolving language patterns.

## II. Literature Review

This section reviews prior work on NER in healthcare and substance use domains, highlighting the techniques, datasets, and evaluation strategies employed, and identifying gaps that motivate the development of specialized approaches such as those explored in the ONER-2025 study.

Miftahutdinov et al. [24] evaluated the effectiveness of multilingual BERT-based models for biomedical named entity recognition (bi) across two languages (English and Russian) and two domains (clinical data and user-generated drug therapy texts). They explore transfer learning (TL) strategies to reduce the need for manually annotated data. The results show that multi-BERT performs best in zero-shot settings when training and test sets are in the same language or domain. TL significantly reduces the labeled data required, achieving 98-99% performance after training on just 10-25% of the sentences.

Ge et al. [25] introduces Reddit-Impacts, a NER dataset focusing on the clinical and social impacts of substance use disorders (SUDs), curated from Reddit discussions about opioids and related medications. The dataset includes manually annotated text spans reflecting personal experiences of nonmedical substance use. The authors aim to create a resource for automatic detection of these impacts from social media data. They applied machine learning models, including BERT, RoBERTa, DANN, and GPT-3.5, to establish baseline performance. The dataset is made available for future research through the 2024 SMM4H shared tasks.

Bose et al. [26] surveys Named Entity Recognition (NER) and Relationship Extraction (RE) techniques in the clinical domain. It reviews existing NLP models, their performance, and challenges in clinical text information extraction. The paper discusses the applications, evaluation metrics, and future research directions. It is the first attempt to address both NER and RE together in the clinical context. The authors highlight the state-of-the-art practices and identify key research articles.

Scepanovic et al. [27] demonstrates how to accurately extract a wide variety of medical entities, such as symptoms, diseases, and drug names, from social media, particularly Reddit. The authors used deep learning with contextual embeddings to outperform state-of-the-art methods on two benchmark datasets (AskaPatient and Micromed). They also created a new benchmark dataset, MedRed, by annotating medical entities in 2,000 Reddit posts. The method was then applied to half a million Reddit posts, categorizing them into 18 diseases, with an average F1 score of 0.87. These results show the potential for cost-effective health behavior modeling and tracking at scale.

Polignano et al. [28] focuses on Named Entity Recognition (NER) in medical documents, an area of growing research due to its social relevance and the challenges posed by short, specific documents. The authors developed a hybrid approach using deep neural networks, comparing various transformer architectures such as BERT, RoBERTa, and ELECTRA to identify the most effective model for their goals. The best-performing model was used in the SpRadIE task of the annual CLEF conference. The results are promising and could serve as a reference for future studies in medical NER.

Wang et al. [29] proposed a new approach for NER called LLMCC, which is designed to improve interactions between different large language models (LLMs). Their study proposes two new approaches, SemnRank and InforLaw-thought, to reduce redundancy in demonstrations and enhance prompt quality. The framework is trained through entity-aware contrastive learning, and extensive experiments show that LLMCC outperforms ten recent studies by over 5% in F1 score across five domains. The research offers valuable domain into information laws, prompting strategies, demonstration selections, and training designs, significantly advancing the use of LLMs in constructing knowledge graphs.

Despite the growing interest in applying Named Entity Recognition (NER) to biomedical and clinical domains, most existing studies primarily focus on structured medical records or formal clinical narratives. There remains a significant research gap in harnessing social media platforms—particularly Reddit, known for its open, in-depth discourse—as a source for extracting opioid-related entities. While prior work has explored entity recognition in Reddit discussions, it either lacks domain-specific focus on opioid-related health discourse or fails to capture the rich context, ambiguity, and informal language unique to user-generated content.

Several studies (e.g., Miftahutdinov et al. [24], Scepanovic et al. [27]) have evaluated multilingual and domain transfer techniques for biomedical NER, yet they do not address the idiosyncrasies of opioid-related conversations, such as slang, emotionally charged expressions, or fragmented posts. Similarly, datasets like Reddit-Impacts (Ge et al.) provide valuable annotated corpora but do not offer fine-grained classification of opioid-related entities like dosage, route of administration, and symptoms.

This study introduces ONER-2025, a novel, manually annotated dataset specifically curated from Reddit posts and labeled across eight domain-specific categories, including drug names, dosage, route of administration, symptoms, Temporal, Location, Event and other. Unlike existing corpora, ONER-2025 addresses the challenges of ambiguous terminology, fragmented sentence structures, and domain-specific slang. Furthermore, this work leverages a comprehensive evaluation using classical machine learning, deep learning architectures (e.g., BiLSTM, CNN), and state-of-the-art transformer models (BioBERT, ClinicalBERT), with pre-trained embeddings like FastText and GloVe to enhance contextual understanding.

Table 1 shows the Comparison of ONER-2025 with related biomedical and social media-based NER datasets. While existing resources like Reddit-Impacts and METS-CoV capture broad clinical or social aspects of substance use or COVID-related health discussions, ONER-2025 introduces a more fine-grained and domain-specific annotation scheme focused on opioid-related entities, including dosage, route of administration, and symptoms. The dataset is uniquely characterized by its handling of ambiguous, slang-filled, and fragmented Reddit language, making it a valuable resource for challenging NER tasks.

TABLE I

COMPARISON OF ONER-2025 WITH RELATED BIOMEDICAL AND SOCIAL MEDIA-BASED NER DATASETS

| Dataset | Source | Entity Types | Annotation Granularity | Language Characteristics | Ref. |
| --- | --- | --- | --- | --- | --- |
| METS-CoV | Twitter (COVID-19) | Disease, Drug, Symptom, Vaccine, Person, Location | Moderate | Short, structured posts, formal-medical language | [30] |
| Reddit-Impacts | Reddit (opioid subreddits) | Clinical Impacts, Social Impacts | Coarse (2 categories) | Long posts, informal, moderately ambiguous | [25] |
| BioNNE | PubMed abstracts | Anatomy, Chemicals, Diseases, Procedures, Devices | High (biomedical taxonomy) | Highly formal, scientific vocabulary | [31] |
| ONER-2025 (Proposed) | Reddit (opioid discussions) | Drugs-Name, Dosage, Route of Administration, Symptoms, Temporal, Event, Location, and Others | Fine-grained, domain-specific | Informal, slang-heavy, ambiguous, fragmented | Proposed |

## III. METHODOLOGY AND DESIGN

### A. Construction of dataset

To build a high-quality dataset of opioid-related discussions from Reddit, we developed a custom Python script specifically tailored for data extraction. This script used the Python Reddit API Wrapper (PRAW) to connect to Reddit's official API. After registering a Reddit application, we authenticated the script using the required credentials, including client ID, client secret, and user agent string. Once connected, we carefully selected five subreddits known for their relevance to opioid use and recovery: r/opiates, r/OpiatesRecovery, r/Drugs, r/Heroin, and r/ChronicPain. These communities were chosen to capture a broad spectrum of user behaviors, ranging from active substance use and illicit sales to personal recovery stories and harm reduction discussions.

To guide the data collection process, we constructed an extensive keyword dictionary that covered a wide range of terminology associated with opioid use. This included not only formal terms such as "opioid," "heroin," "oxycodone," "fentanyl, injecting heroin, snorting opioids, intranasal fentanyl, smoking heroin, and smoking fentanyl" but also an extensive list of slang words (e.g., "fent," "roxie," "perk," "black," "chocolate"), street names, and commonly observed misspellings (e.g., "herone," "oxycotin," "fentanil" etc.). Including these informal expressions was essential to ensure comprehensive coverage of real-world language used in online drug discourse. The script was designed to collect both original and top-level posts, capturing critical metadata such as post title, body content, timestamp, subreddit name, and anonymized usernames. We implemented additional filters to clean the data, remove empty or irrelevant posts, and avoid duplication. To ensure ethical data collection and comply with Reddit's rate limits, our script included timed delays and logged extraction checkpoints. The data was gathered over a 3 years, from January 2021 to December 2023, ensuring temporal diversity and the inclusion of both long-term and short-term trends in opioid-related conversations. Once collected, the dataset was saved in Excel (.xlsx) format for further annotation and analysis. This process resulted in a rich and nuanced dataset that reflects real user behavior and language surrounding opioids on social media—forming the foundation for the multi-class classification and NLP experiments that followed. Figure 1 shows the overall Architecture of proposed methodology and design of the study.

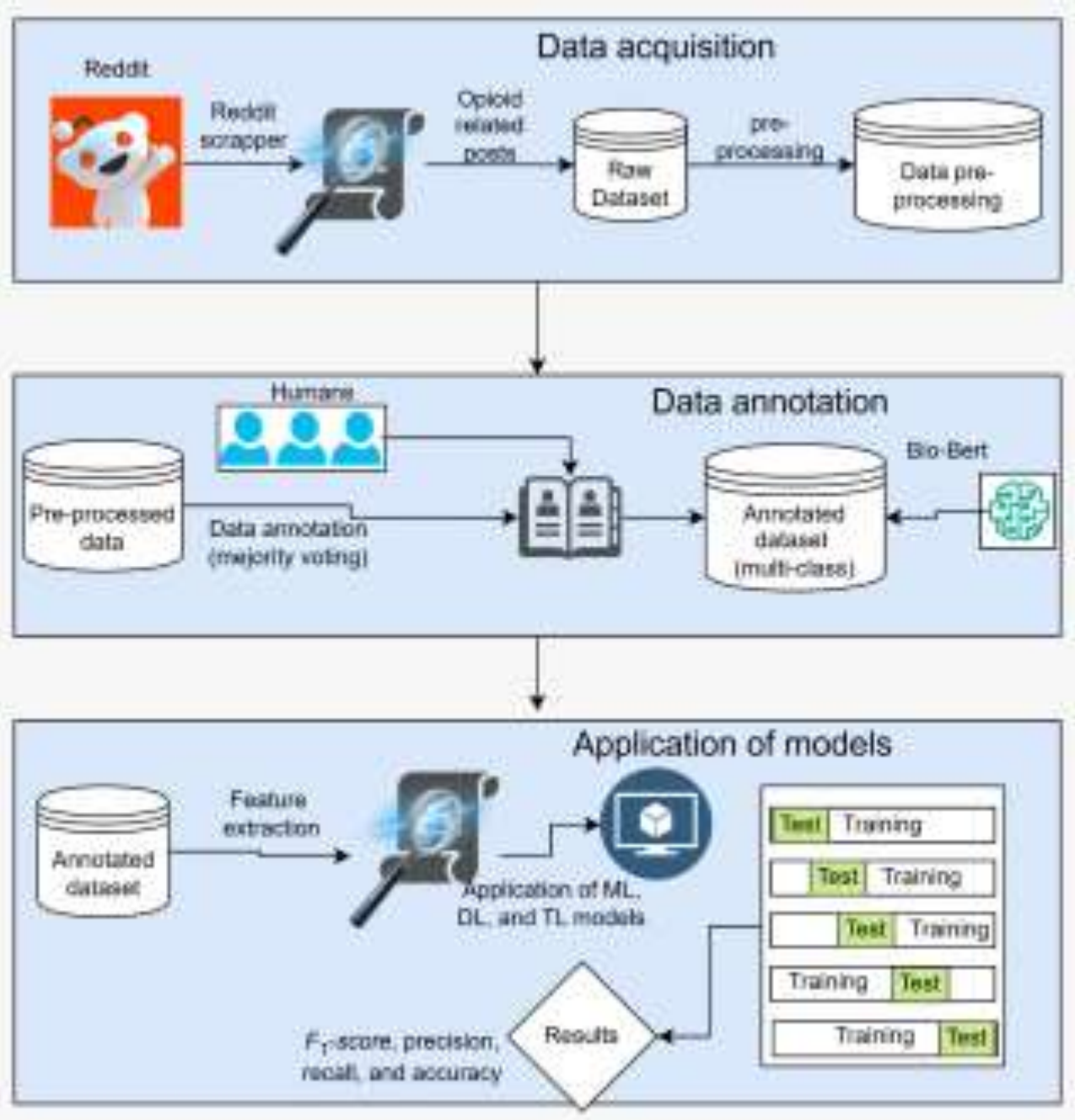

Fig 1. Architecture of proposed methodology.

## B. Pre-processing

Preprocessing plays a critical role in preparing the NER dataset for training machine learning models, particularly when working with real-world social media data that is often noisy and multilingual. As illustrated in Figure 2, we applied a structured preprocessing pipeline to the raw Reddit posts related to opioid use to ensure data quality and model-readiness. First, we removed irrelevant or noisy components such as hashtags, emojis, and hyperlinks using Python's re module for regular expressions and the emoji library for emoji filtering. The text was then normalized by converting all characters to lowercase to ensure consistency. We further employed a custom lexicon and the contractions library to expand shortened forms, and abbreviations into their complete forms (e.g., "u" → "you", "can't" → "cannot"), improving the model's contextual understanding. Posts containing fewer than 20 characters were filtered out, as they typically lacked meaningful content for downstream entity recognition. This preprocessing strategy ensures the resulting dataset is clean, semantically rich, and well-suited for training robust opioid-related NER models.

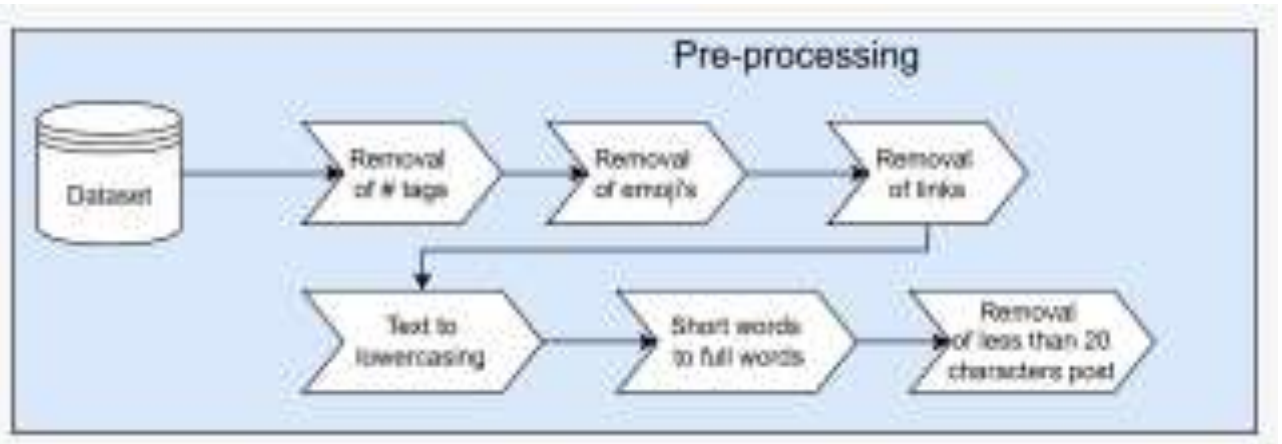

Fig 2. Pre-processing steps utilized in the study.

## C. Annotation guidelines

The annotation guidelines for ONER-2025 focus on accurately identifying and categorizing opioid-related entities from social media discourse. These guidelines were developed in close collaboration with a domain expert to ensure clinical relevance and clarity in handling complex, real-world expressions of opioid use. Annotators are instructed to label eight predefined categories: Drug-Name, Dosage, Route of Administration, Symptoms, Temporal, Location, Event, and Others. Special attention is given to slang terms (e.g., *"blues"* for oxycodone), ambiguous words, and fragmented or informal sentences common in online discussions. Contextual understanding is essential to distinguish between different meanings and ensure precise entity recognition, thereby enabling robust and accurate opioid-related named entity recognition. The ONER-2025 dataset includes the following guidelines to annotate the entity categories:

- Drugs-Name: Label all mentions of opioid substances and related drugs, including both standard (generic or brand) names and slang terms. Examples include: fentanyl, fent, Roxicodone, oxycodone, heroin, morphine, perc, oxy, and roxy. All such references should be annotated as Drug-Name. For example, the phrase "He got hooked on the green monster" was annotated as 'green monster → heroin (Drug-Name)', based on community-sourced slang references and contextual cues indicating drug dependence.
- Dosage: Label all mentions of the quantity or amount of drug intake, expressed in either numerical or textual form. Examples include: 10 mg, two pills, and half a tab, take half a pill, one shot, and one dose. A challenging case involved "just needed a bump," where "bump"—a slang term referring to a small dose of powdered drug—was annotated as Dosage based on its pairing with known opioid slang.
- Route of Administration: To annotate the route of administration, annotators must identify the described method by which a drug is consumed by the opioid user. For example: oral, snorting, injecting, and intranasal.
  In informal tweets like "sniffed a bit before class," the term "sniffed" was recognized as slang for intranasal drug use and annotated accordingly.
- Symptoms: To annotate the label as symptoms annotator must read the full sentence and captures physical or psychological effects associated with opioid drug abuse. For example respiratory depression, unconsciousness, nausea, euphoria, dehydration, mood swings, aggression, suicidal thoughts.

- Temporal: Label all expressions that refer to time, duration, or specific time intervals related to opioid use. This includes both explicit and relative temporal references such as *today*, *yesterday*, *last night*, *next week*, *within an hour*, *within a day*, and any mentions indicating past or future drug intake. These should be annotated as Temporal.
- Location: To annotate the Location as label annotator must read the sentence and Identify a geographic locations, Addresses and Landmarks or Geopolitical Terms where opioid-related activities occur. For example New York, hospital, rehab center, street corner, national borders, international boundaries.
- Event: To annotate the Event as a Label annotator must identify the denoted significant incidents or occurrences related to opioid use, for example overdose, prescription refill, drug bust, rehab admission, hospital admission, and emergency response.
- Others: to annotate the other class annotator must captures opioid-related entities that do not fit into the above categories but still provide relevant context for understanding the opioid-related discussion. For example addiction, withdrawal, dependency, recovery, treatment, cravings etc.

### D. Handling Ambiguity

- Slang Usage: To handle slang terms used in a sentence, annotators must read the sentence carefully. If a term is commonly used as slang for an opioid, it should be annotated under Drug-Name. For example: "Looking for blues tonight" blues (Drug-Name). Another example includes "Need some fire," where "fire" was interpreted as high-potency heroin after examining user history and slang databases.
- Ambiguity (Multiple Meanings): If a term has multiple interpretations, context should determine the annotation. For example: "Need some oxy ASAP" → oxy (Drug-Name) or oxygen (Other) depending on sentence context. In one tweet, "he ODed on H"—the letter 'H' could mean heroin or something else, but due to proximity with 'ODed' (overdosed), it was annotated as heroin (Drug-Name).
- Fragmented Sentences**:** In cases where an opioid-related entity is implied rather than explicitly stated, it should be annotated if contextually clear. For example: "Got some percs... took too many earlier today, feeling a little off." Percs (Drug-Name). Similarly, "just popped 2 before work" was annotated with "popped" implying ingestion, and '2' understood as dosage—annotated as Dosage.

### E. Annotation Process

Annotation is the process of identifying named entities from unstructured text and classifying them into predefined entity types, which helps improve model performance in supervised learning methods, especially for complex tasks. In this study, we employed a manually annotated dataset using a hybrid approach that combined automated labeling with manual verification to ensure both efficiency and accuracy. The annotation process was carried out in collaboration with two human expert annotators and the first author, who has prior experience in opioid data annotation. Additionally, we utilized a variant of pre-trained BERT models, such as BioBERT, to support the annotation process. Bio-BERT is trained on approximately 13.5 billion words from large biomedical corpora to identify diseases, drugs, genes and proteins, chemical compounds, anatomical entities, medical procedures, biomedical events, and clinical terms. Below is the detailed procedure followed.

- Initial human based annotation: Initially, we strictly followed defined annotation guidelines and manually annotated 2,000 Reddit posts related to opioid use for NER. This phase focused on identifying entities such as Drugs-Name, Dosage, Route of administration, Symptoms, Temporal, Location, Event, and Others. These high-quality, human-labeled samples served as the foundation for fine-tuning the BioBERT model.
- Model-Assisted Annotation: After training BioBERT on the manually annotated dataset, we used the fine-tuned model to automatically annotate an additional 2,000 samples. The model generated entity labels with a confidence threshold set at 90% to ensure quality. This automated process provided initial annotations for each text instance, targeting the same entity types, and significantly accelerated the annotation pipeline while maintaining accuracy.
- Human based-Annotation correction: after the model-assisted annotations two human annotators were employed to manually review the output generated by the model. Both annotator individually observed the labeled entities, making adjustments as needed to correct any errors. For example, If the model misidentified an entity (e.g., marking a symptom as a drug name, temporal expressions as dosage entities, snorted" as a drug name instead of recognizing it as a route of administration), the annotator adjusted the label accordingly.

This procedure helped us to ensure both efficiency and accuracy the model's accuracy by providing high-quality, manually-verified annotations. With these refined annotations, we retrained the model on 4000 annotated samples and had it annotate another 2000 samples, again maintaining the 90% threshold for accuracy and we repeated step 2 again. Finally, we fine-tuned the model on 6357 samples, iteratively improving its performance. This structured approach, combining model automation with human expertise, allowed us to develop a high-quality opioid NER dataset with reliable and accurate entity recognition.

During this process, we observed that some model misclassifications—particularly in overlapping or ambiguous contexts, such as confusing drug dosages with temporal

expressions—were consistently corrected by human annotators with majority voting. These observations highlight how human input played a critical role in refining model predictions and reducing systematic labeling errors.

- Discrepancy Resolution: In some cases, where both annotators disagreed on a specific label, a third step was introduced. A meeting was convened in which both annotators and the corresponding author discussed the issue in detail and reached a consensus on the correct label through majority voting.
- Final Review: Once the human annotators reviewed and improve the model's annotations, we conduct a final quality check. Any remaining ambiguities were resolved according to the pre-defined categories.
- Output Dataset: The final dataset consisted of fully 8357 annotated samples containing 331,285 tokens and 8 unique categories for training the ONER-2025 models.

By employing this hybrid data annotation approach the procedure facilitated us for the creation of a high-quality labeled dataset named as ONER-2025 dataset. Table 2 shows the recommended parameters used for fine-tuning BioBERT on Opioid named entity recognition, we employed the BioBERT model with the AdamW optimizer and a learning rate of 2e-5, along with a weight decay of 0.01 to prevent over-fitting. Gradient clipping was set to 1.0 to stabilize training. The model was trained with a batch size of 16 for 10 epochs, using a maximum sequence length of 128 and a dropout rate of 0.1 to improve generalization. To regulate learning, we implemented a linear decay scheduler with warm-up, and early stopping was applied if no improvement was observed within 2-3 epochs. The cross-entropy loss function was used to optimize classification performance, ensuring robust and accurate entity recognition.

TABLE II

BIO-BERT FINE TUNING PARAMETER DURING THE DATA ANNOTATION PHASE

| Category | Parameter | Recommended Value |
|---|---|---|
| Model & Optimizer | Model | Bio BERT |
| | Optimizer | AdamW |
| | Learning Rate | 2e-5 |
| | Weight Decay | 0.01 |
| | Gradient Clipping | 1.0 |
| Training Parameters | Batch Size | 16 |
| | Epochs | 10 |
| | Max Sequence Length | 128 |
| | Dropout Rate | 0.1 |
| Regularization &Learning Strategy | Scheduler | Linear Decay with Warmup |
| | Early Stopping | Stop if no improvement in 2-3 epochs |
| | Loss Function | Cross-Entropy Loss |

## F. Inter-Annotator Agreement

During the annotation process, occasional disagreements arose among annotators due to the informal, ambiguous, and slang-heavy nature of Reddit discourse. To quantify annotation reliability, we calculated Cohen's Kappa coefficient, which yielded a value of 0.80 — indicating substantial agreement and demonstrating the strength of our annotation guidelines. To resolve annotation disagreements, we employed a structured adjudication process. In cases of conflict, annotations were collectively reviewed and resolved through a majority voting system. This expert involvement helped ensure the clinical accuracy and contextual appropriateness of the final labels. Table 3 provides the interpretation scale for Cohen's Kappa, supporting the conclusion that the annotation process achieved high inter-rater reliability.

TABLE III

SHOWS THE INTERPRETATION OF KAPPA VALUES

| Kappa Value | Interpretation |
|---|---|
| 0.00 - 0.20 | Poor Agreement |
| 0.21 - 0.40 | Fair Agreement |
| 0.41 - 0.60 | Moderate Agreement |
| 0.61 - 0.80 | Substantial Agreement |
| 0.81 - 1.00 | Almost Perfect Agreement |

## G. Corpus Characteristics and Standardization

The table 4 provides a comprehensive overview of the dataset's characteristics. The dataset contains a total of 8,357 samples distributed across 8 distinct classes. It includes 331,285 tokens, with a vocabulary size of 15,867 unique words. On average, each sentence in the dataset contains approximately 39.64 tokens, with the shortest sentence having 3 tokens and the longest containing 92 tokens. Additionally, the dataset features a total of 319,588 labeled entities, indicating the volume of labeled data available for potential tasks such as fine-tuning or model training. Table 5 shows the label distribution of our ONER-2025 datasets, while Figure 3 (word cloud) provides a quick visual summary of the most frequent words in a dataset. It helps readers identify common themes or key terms without reading large amounts of text.

TABLE IV

STATISTICS OF ONER-2025 DATASETS

| Characteristic | Value |
|---|---|
| Total Samples | 8357 |
| Total Classes | 8 |
| Total Tokens | 331285 |
| Vocabulary Size | 15867 |
| Average Tokens per Sentence | 39.64 |
| Max Tokens in a Sentence | 92 |
| Min Tokens in a Sentence | 3 |
| Total Entities | 319,588 |

TABLE V

LABEL DISTRIBUTION OF DATASET

| Entity Type | Number of Tokens |
|---|---|
| Drug_name | 15,145 |
| ROA | 4,242 |
| Event | 7,611 |
| Location | 2,414 |
| Symptom | 4,864 |
| Other | 273,165 |
| Dosage | 2,709 |
| Temporal | 9,438 |

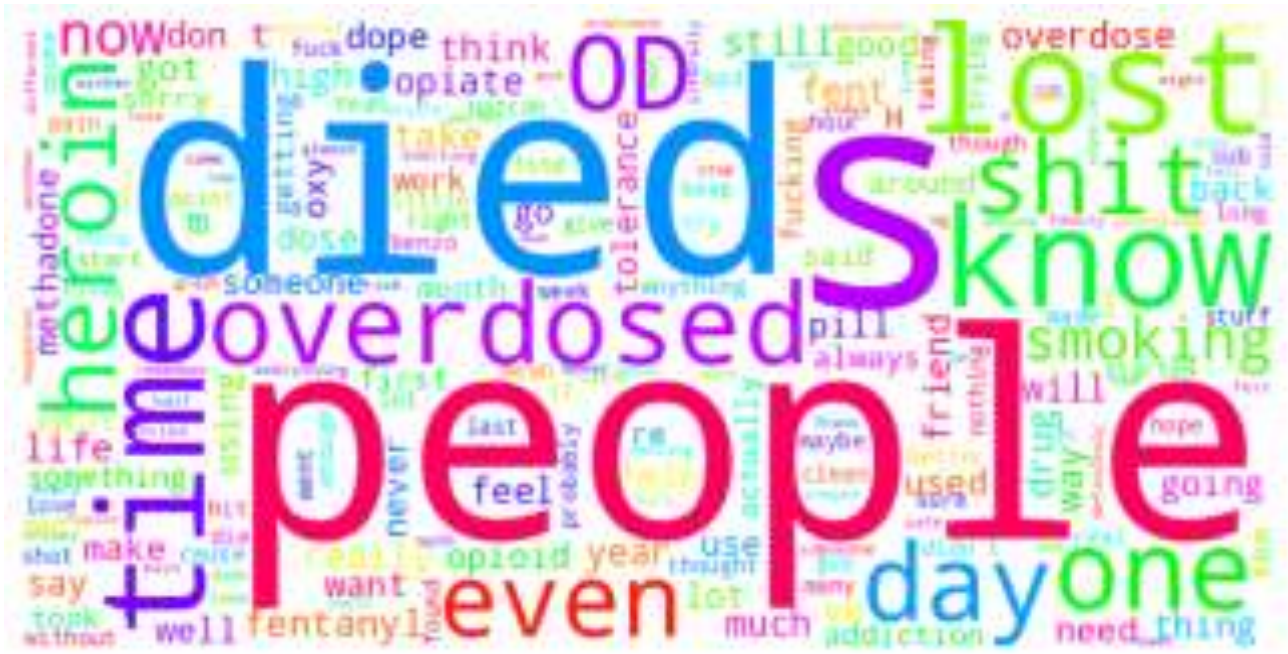

Fig 3. Word cloud representation of the ONER-2025 dataset.

### H. *Ethical Statement*

This study exclusively utilized publicly available, anonymized data from Reddit, a platform where users post voluntarily and often under pseudonyms. No personally identifiable information (PII) was collected or stored during the data collection or annotation process. The research protocol complied with Reddit's Terms of Service and relevant data usage policies. Given that no human subjects were directly involved and no intervention was conducted. The study was conducted solely for academic research purposes with a focus on public health surveillance. Efforts were made to minimize any potential risks of re-identification or misuse of data.

### I. *Application of models Training and testing phase*

In this phase, we present a structured pipeline for applying machine learning models—including Support Vector Machine (SVM), Decision Tree (DT), and Logistic Regression (LR)—using token-based feature extraction methods. We employed handcrafted token-level features, explicitly avoiding TF-IDF or bag-of-words, which are not suitable for sequence labeling tasks such as Named Entity Recognition (NER). Additionally, we implemented two deep learning models—BiLSTM and CNN—using FastText and GloVe word embeddings, as well as transformer-based language models including BERT-base-NER, RoBERTa-base, BioBERT-base-cased, and Bio_ClinicalBERT, which leverage advanced contextual language embedding techniques. The reason for using machine learning, deep learning, and transformer models together in a single task to check the strength of each approach to achieve optimal performance and find the best model for more accurate results for a specific ONER-2025 task. Training and testing process for machine learning, deep learning, and language-based models, illustrating the standard approach to predictive model development. The process begins with a comprehensive ONER-2025 dataset, which serves as the foundation for learning. This dataset is divided into 5-fold cross validation for training and testing set as shown in figure 4. The training set is directed into the learning phase, where models based architectures are developed. These models utilize the training data to learn patterns and relationships within the dataset by optimizing parameters through iterative processes.

Simultaneously, the testing set is reserved for later use and remains untouched during the training phase to ensure an unbiased evaluation of the model's performance. Once the training is complete, the resulting model enters the prediction phase. Here, the model processes the testing set, generating predictions based on its learned parameters as shown in Table 6, 8 and 10. This iterative approach ensures that the model not only fits the training data but also performs well on unseen data, validating its robustness and reliability.

In the model evaluation phase, four popular performance metrics such as precision, recall, F1-score and cross validation score were used to assess performance of the models. These metrics are critical in understanding how well the model performs across various aspects of prediction accuracy, robustness, and reliability. The key metrics used are cross validation score, precision, recall, and F1-score are given in the following equations.

$$\text{CV Score} = \frac{1}{k}\sum_{i=1}^{k} Score_i \quad (1)$$

$$\text{Precision} = \frac{TP}{FP + TP} \quad (2)$$

$$\text{Recall} = \frac{TP}{FN + TP} \quad (3)$$

$$\text{F1 Score} = 2 \times \frac{\text{Recall} \times \text{Precision}}{\text{Recall} + \text{Precision}} \quad (4)$$

While k is the number of folds in cross-validation, $Score_i$ is the performance score for fold i, and TP is True Positive, TN is True Negative, FP is False Positive, and FN is False Negative.

By analyzing these metrics together, a comprehensive understanding of the model's strengths and weaknesses is achieved. A robust model demonstrates high accuracy while maintaining a balance between precision, recall, and F1-score, ensuring reliable performance across diverse datasets and problem domains.

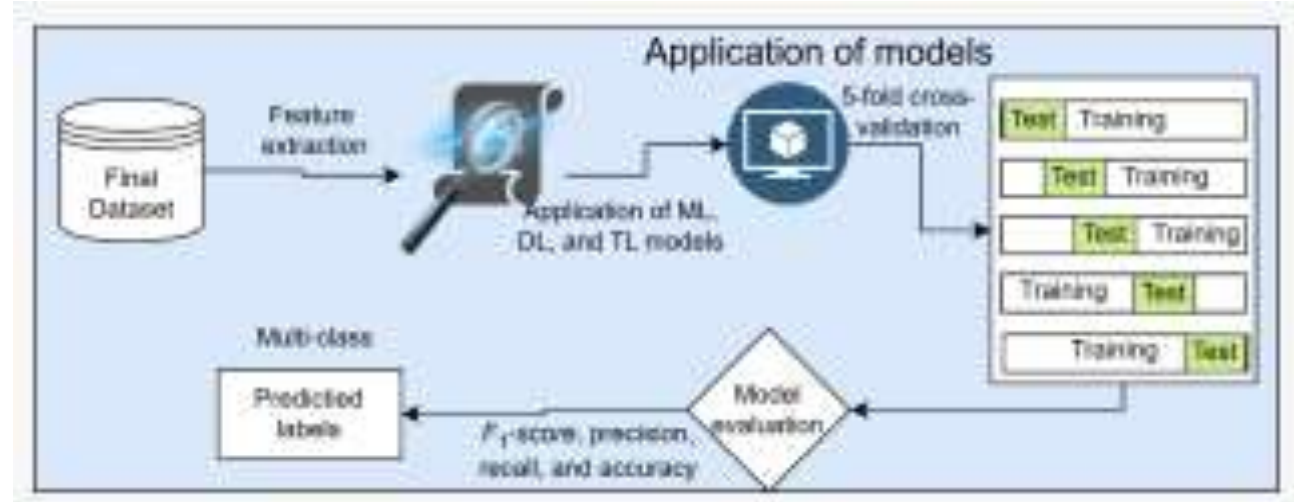

Fig 4. Application of Models, Training and testing phase Word.

## IV. Results

In this section, we present and analyze the results obtained from applying various machine learning, deep learning, and advanced transformer-based techniques, as described in the previous methodology section. The objective of this analysis is to systematically evaluate and compare the performance of these approaches in addressing the Opioid Named Entity Recognition (ONER) task. Traditional machine learning models such as Random Forest (RF), Support Vector Machine (SVM), and Logistic Regression were first employed as baseline methods to establish a foundational performance benchmark. These were followed by deep learning architectures, including Bidirectional Long Short-Term Memory (BiLSTM) and Convolutional Neural Networks (CNN), which are well-known for their ability to capture contextual patterns in sequence labeling tasks.

To further improve entity recognition in complex and noisy social media texts, we implemented state-of-the-art transformer-based models, including BERT-base and RoBERTa-base. These models are pre-trained on large corpora and fine-tuned on our ONER-2025 dataset to extract contextual and domain-specific features effectively. The performance of each model was evaluated using standard metrics such as accuracy, precision, recall, and F1-score, across 5-fold cross-validation settings. Through this comparative analysis, we aimed to identify the most robust and generalizable model capable of accurately detecting opioid-related entities in unstructured Reddit discussions. The following subsections provide a detailed breakdown of the results, highlighting the relative strengths and limitations of each modeling approach.

## A. Results for Machine learning

The table 6 summarizes the best-performing settings found for three machine learning models used in the opioid Named Entity Recognition (NER) project. For Logistic Regression, the model worked best with a regularization strength of C=1.0, an L2 penalty to prevent over fitting, and the lbfgs solver for optimization. The Support Vector Machine gave optimal results with a stronger regularization parameter C=10, using an RBF kernel to handle non-linear relationships, and a gamma value of 0.01 to control the influence of each data point. For the Random Forest, the best configuration included 100 decision trees (n_estimators=100), a maximum depth of 10 to balance complexity and generalization, and a minimum split size of 2 samples to create new branches. In simpler terms, these tuned settings allowed each model to best identify and classify opioid-related entities from the data while avoiding overfitting and maintaining accuracy.

TABLE VI

HYPER-PARAMETER OF TUNING OF MACHINE LEARNING MODELS USED IN THIS STUDY

| Model | Hyperparameters | Optimal Values Found |
|---|---|---|
| Logistic Regression (LR) | C, penalty, solver | C=1.0, penalty='l2', solver='lbfgs' |
| Support Vector Machine (SVM) | C, kernel, gamma | C=10, kernel='rbf', gamma=0.01 |
| Random Forest (RF) | n_estimators, max_depth, min_samples_split | n_estimators=100, max_depth=10, min_samples_split=2 |

The table 7 shows the performance evaluation of three different machine learning models including Logistic Regression (LR), Support Vector Machine (SVM), and Random Forest (RF)—across four key evaluation metrics such as Precision, Recall, F1-score, and Accuracy. All models achieve a recall of 0.88, demonstrating their capability to correctly classify positive instances effectively. While LR and SVM have equal precision (0.83), F1-score (0.83), and accuracy (0.88), signifying similar performance in handling both false positives and false negatives. However, RF perform best with a slightly higher precision=0.86), meaning it makes fewer false-positive predictions compared to the other two models. Despite this, its F1-score remains the same as the others (0.83), indicating a balanced trade-off between precision and recall. The overall accuracy of machine learning models is 0.88, reflecting their effectiveness in making correct predictions. This assessment shows that while all models perform well but RF is slightly better at reducing false positives.

TABLE VII

RESULTS FOR MACHINE LEARNING MODELS

| Model | Precision | Recall | F1-score | C.V score |
|---|---|---|---|---|
| LR | 0.83 | 0.88 | 0.83 | 0.88 |
| SVM | 0.83 | 0.88 | 0.83 | 0.88 |
| RF | 0.86 | 0.88 | 0.83 | 0.88 |

## B. Results for Deep Learning

The table 8 shows the hyper parameters of deep learning models such as BiLSTM (Bidirectional LSTM) and CNN (Convolutional Neural Network), along with their corresponding values for a grid search process. In both models, the learning rate of 0.1 is used, 5 epochs, Embedding dimension 300, and batch size 32 are used. BiLSTM model takes 128 units as its LSTM layers because this enables the model to learn sequential patterns in both directions. Conversely, the CNN model contains 128 filters in its convolutional layers, and every filter has its own kernel size of 5, which assist it in the extraction of features on the basis of an input data by means of a sliding window. This grid search setup aims to optimize these hyper parameters for each model to enhance performance on the given task.

TABLE VIII

HYPER-PARAMETER OF TUNING OF DEEP LEARNING MODELS USED IN THIS STUDY

| Model | Hyperparameters | Grid Search Values |
|---|---|---|
| BiLSTM | Learning Rate | 0.1 |
| | Epochs | 5 |
| | Embedding Dim | 300 |
| | Batch Size | 32 |
| | LSTM Units | 128 |
| CNN | Learning Rate | 0.1 |
| | Epochs | 5 |
| | Embedding Dim | 300 |
| | Batch Size | 32 |
| | Filters | 128 |
| | Kernel Size | 5 |

The table 9 compares the performance of two models, CNN and BiLSTM, using two different word embeddings: GloVe and FastText. The metrics shown—Precision, Recall, F1-score, and Accuracy—measure how well each model performs in terms of classification. For GloVe embeddings, the BiLSTM model outperforms the CNN model in all metrics, with an F1-score of 0.89 and accuracy of 0.88, while the CNN scores 0.85 for F1 and 0.86 for accuracy. For FastText embeddings, BiLSTM still leads in Precision and F1-score (0.88 and 0.87), but the CNN model has a slightly higher Recall of 0.87, and its overall accuracy (0.87) is slightly better than BiLSTM's (0.88). Overall, BiLSTM generally performs better with GloVe embeddings, while the models are closer in performance with FastText embeddings.

TABLE IX

RESULTS FOR DEEP LEARNING MODELS

| Model | Precision | Recall | F1-score | Accuracy |
|---|---|---|---|---|
| GloVe | | | | |
| CNN | 0.85 | 0.86 | 0.85 | 0.86 |
| BiLSTM | 0.89 | 0.88 | 0.89 | 0.88 |
| FastText | | | | |

| CNN | 0.85 | 0.87 | 0.85 | 0.87 |
|---|---|---|---|---|
| BiLSTM | 0.88 | 0.88 | 0.87 | 0.88 |

## C. Transformer Result

The table 10 outlines the hyperparameters for four different transformer-based models such as bert-base-NER, roberta-base, biobert-base-cased, and Bio_ClinicalBERT—with their associated values for a specific training setup. Each model is trained for 5 epochs, with a learning rate of 5e-5, indicating a relatively small update to the model weights during training. The batch size is set to 16, meaning 16 examples are processed at once before updating the model's weights. The max length for input sequences is capped at 128 tokens, ensuring that inputs longer than this are truncated, which is typical for models dealing with textual data. The optimizer used is AdamW, a variant of the Adam optimizer that helps with weight decay, making it suitable for training large transformer models like these.

TABLE X

HYPER-PARAMETER OF TUNING OF TRANSFER LEARNING MODELS USED IN THIS STUDY

| Model Name | Hyperparameter | Value |
|---|---|---|
| bert-base-NER, roberta-base, biobert-base-cased, Bio_ClinicalBERT | Epochs | 5 |
| | Learning Rate | 5e-5 |
| | Batch Size | 16 |
| | Max Length | 128 |
| | Optimizer | AdamW |

The table 11 presents the performance metrics of four different machine learning models used for named entity recognition (NER). Each model's effectiveness is evaluated using precision, recall, F1-score, and accuracy. The precision, recall, and F1-score all range from 0.96 to 0.97 across the models, indicating that they perform consistently well in correctly identifying relevant entities. Precision refers to the proportion of true positives out of all predicted positives, while recall measures the proportion of true positives out of all actual positives. The F1-score is the harmonic mean of precision and recall, balancing the two. Accuracy represents the overall correctness of the model in terms of both true positives and true negatives. All models show strong results, with BERT-based models (bert-base-NER and roberta-base) achieving slightly higher accuracy and scores, suggesting their strong overall performance in the NER task. Bio_ClinicalBERT, while still effective, has slightly lower scores in comparison but remains highly competent in the task.

TABLE XI

RESULTS FOR TRANSFORMER MODELS

| Model | Precision | Recall | F1-score | Accuracy |
|---|---|---|---|---|
| bert-base-NER | 0.97 | 0.97 | 0.97 | 0.97 |
| Roberta-base | 0.97 | 0.97 | 0.97 | 0.97 |
| biobert-base-cased- | 0.97 | 0.96 | 0.97 | 0.96 |
| Bio_Clinical-bert | 0.96 | 0.96 | 0.96 | 0.96 |

## D. Error Analysis

The Table 12 shows the comparative performance of four top-performing models—Random Forest (RF), BiLSTM with GloVe embeddings, BERT-base-NER, and RoBERTa-base—on the task of Opioid Named Entity Recognition (NER), evaluated using Precision, Recall, F1-score, and Accuracy. Transformer-based models, BERT-base-NER and RoBERTa-base, consistently outperform the others across all metrics, each achieving a near-perfect score of 0.97 in every category, demonstrating their superior ability to accurately and reliably detect opioid-related named entities in complex social media text. In contrast, BiLSTM (GloVe) shows moderate performance, with scores ranging from 0.88 to 0.89, indicating it captures some contextual information but lacks the deep contextual understanding of transformer models. Random Forest performs the weakest, especially in F1-score (0.83), highlighting its limited capacity to handle the linguistic variability and nuance present in Reddit discussions. Overall, the chart underscores the effectiveness of transformer-based architectures in domain-specific NER tasks within noisy and informal text environments like social media. While figure 5 shows the Confusion matrix of Bert-base-NER model, figure 6 shows the Confusion matrix of XLM-R model, figure 7 shows the Training and validation performance of different epochs of Bert-base-NER model and figure 8 shows the Training and validation performance of different epochs of XLM-R model.

TABLE XII

TOP PERFORMING MODELS IN EACH LEARNING APPROACH

| Model | Precision | Recall | F1-score | Accuracy |
|---|---|---|---|---|
| RF | 0.86 | 0.88 | 0.83 | 0.88 |
| BiLSTM | 0.89 | 0.88 | 0.89 | 0.88 |
| bert-base-NER | 0.97 | 0.97 | 0.97 | 0.97 |
| Roberta-base | 0.97 | 0.97 | 0.97 | 0.97 |

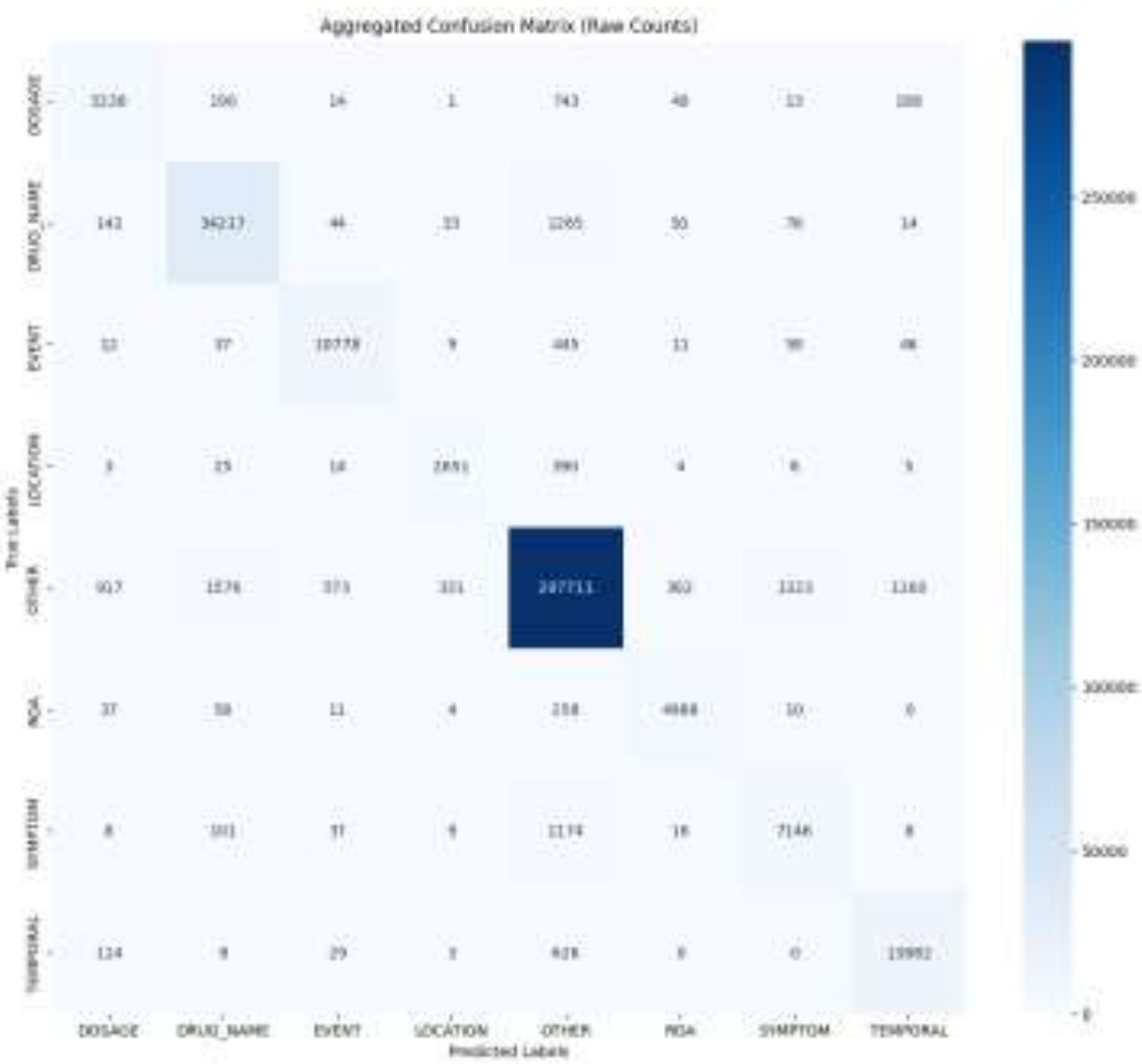

Fig 5. Confusion matrix of Bert-base-NER model.

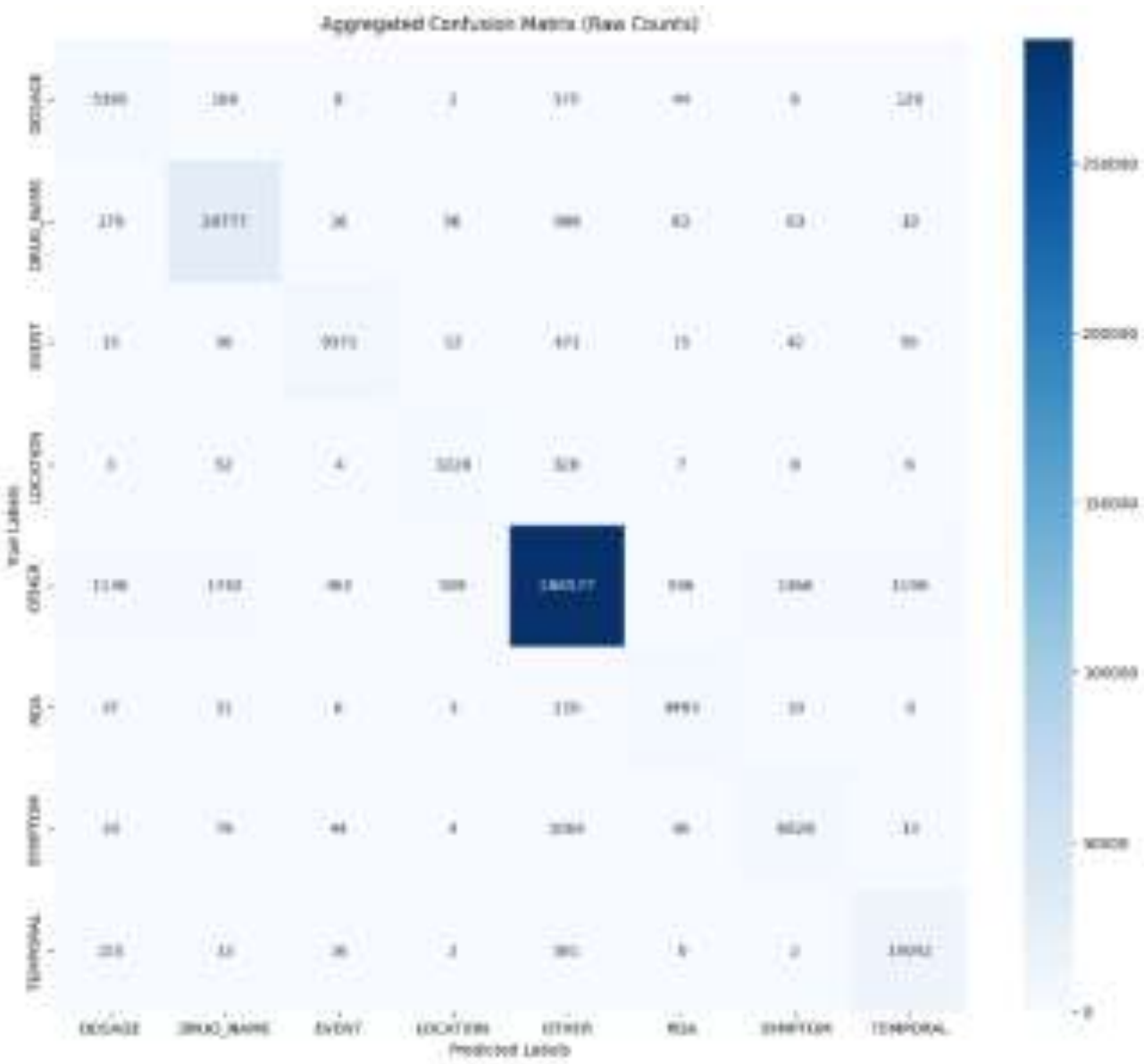

Fig. 6. Confusion matrix of Roberta-base model.

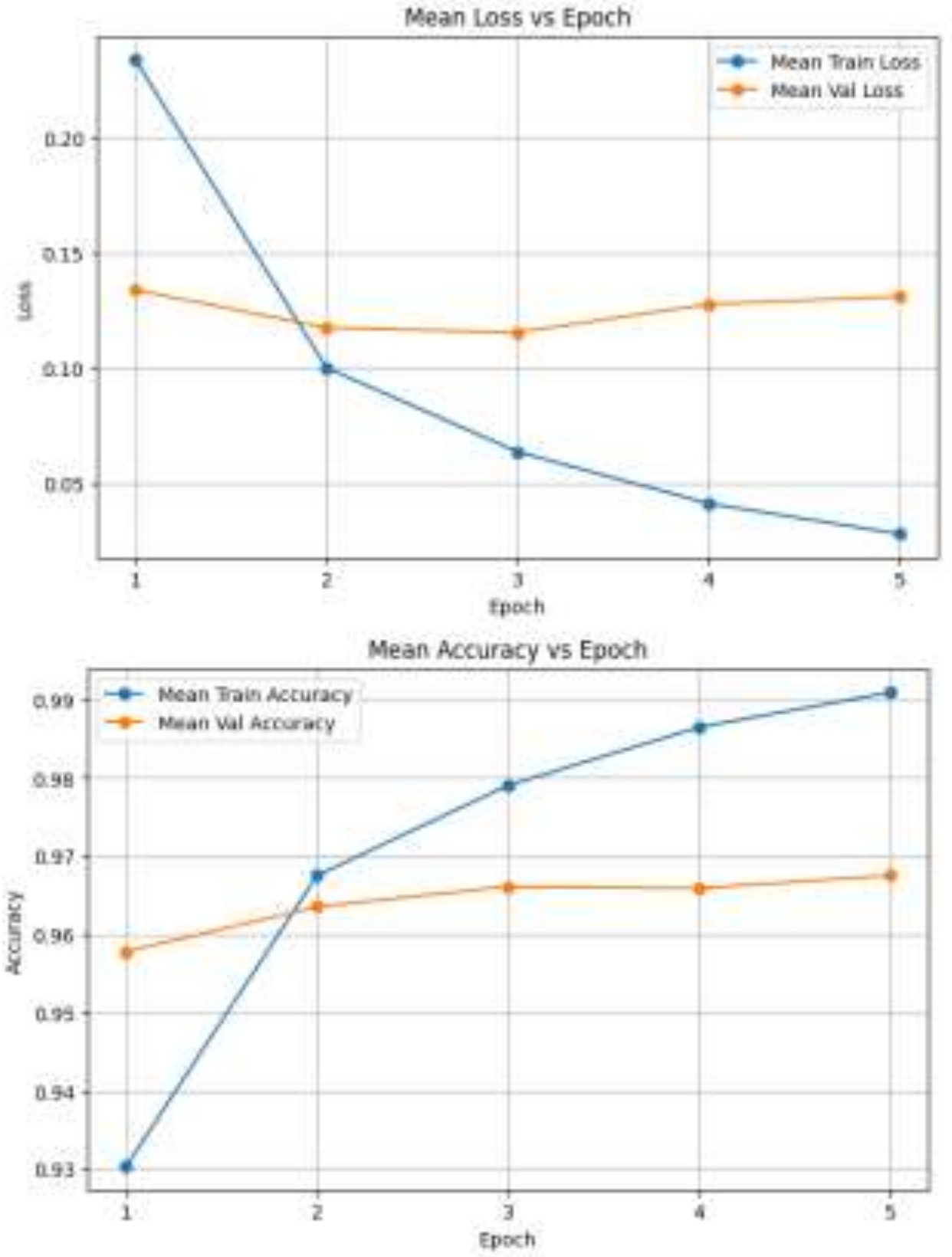

Fig. 7. Training and validation performance of different epochs of Bert-base-NER model.

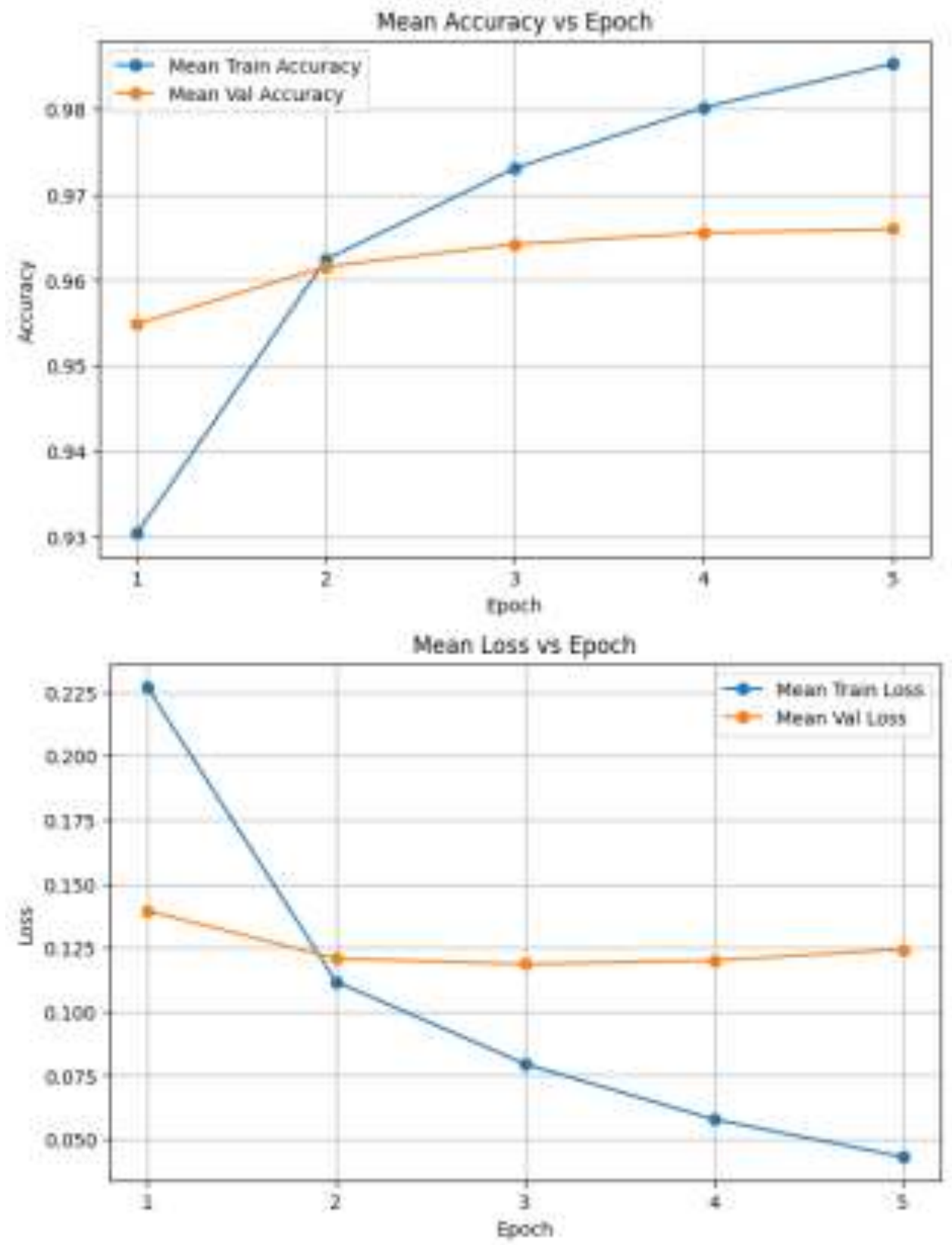

Fig. 8. Training and validation performance of roberta-base model.

### E. Statistical Analysis

Table 13 shows the paired t-test results comparing the performance of different models against the baseline Random Forest (RF) model. The t-test was performed to check whether the improvements in F1-scores by BiLSTM (GloVe), bert-base-NER, and roberta-base were statistically significant or just due to chance.

The results clearly show that bert-base-NER significantly outperforms both Random Forest (RF) and BiLSTM (GloVe). Specifically, the difference between bert-base-NER and RF is highly significant ($p < 0.00000004$), and the difference with BiLSTM (GloVe) is also very strong ($p < 0.000003$). These low p-values mean there's extremely strong evidence that bert-base-NER performs better—not just due to chance, but due to consistent improvement across all folds.

When comparing bert-base-NER to roberta-base, the difference is much smaller. Although bert-base-NER shows slightly higher F1 scores on average, the difference is not statistically significant ($p = 0.066$), meaning their performance is roughly equivalent from a statistical standpoint.

In summary, bert-base-NER is significantly better than RF and BiLSTM (GloVe), and comparable in performance to roberta-base. This highlights how transformer-based models, particularly BERT and RoBERTa, provide a clear advantage over traditional machine learning and even earlier deep learning approaches for Named Entity Recognition (NER) tasks in the context of opioid-related discussions.

TABLE XIII

STATISTICAL COMPARISON OF TOP PERFORMING MODELS USING T-TEST

| Model | f1 | f2 | f3 | f4 | f5 |
|---|---|---|---|---|---|
| RF | 0.8742 | 0.8765 | 0.8769 | 0.8761 | 0.8762 |
| BiLSTM (GloVe) | 0.8902 | 0.8874 | 0.8864 | 0.8843 | 0.8763 |
| bert-base-NER | 0.9686 | 0.9672 | 0.9667 | 0.9681 | 0.9671 |
| Roberta-base | 0.965 | 0.9671 | 0.9659 | 0.9658 | 0.9661 |

### F. Qualitative Analysis

To gain deeper insight into how our top-performing models handle ambiguous and informal social media text, three annotators—the first author along with two annotators—carefully reviewed and analyzed predictions from Random Forest, BiLSTM, and XLM-R on a selection of posts from the ONER-2025 dataset. In the post *"I just got my refill for oxy. Gonna try a smaller dose this time, don't wanna end up like last week,"* RF detected only the drug name "oxy" and missed the event and dosage-related phrases. BiLSTM identified "oxy" and "smaller dose" but failed to associate "refill" with a medication-related event. XLM-R outperformed both by capturing all relevant entities; however, in other posts, even XLM-R showed limitations. For example, in *"Feeling numb again. Popped 20 mg of blues around noon, hope I don't black out,"* XLM-R correctly identified the dosage and slang term "blues" as a drug name, but inconsistently labeled the symptom "black out" in some similar test cases, likely due to contextual ambiguity or limited training instances for such emotional expressions. Another key challenge is recognizing the routes of administration—how individuals consume opioids.

For instance, in the post *"Crushed some percs last night, sniffed half and saved the rest,"* "sniffed" indicates the intranasal route. While RF and BiLSTM often missed this informal term, XLM-R successfully detected it, showing stronger grasp of varied and informal language. While XLM-R clearly demonstrates stronger contextual understanding and robustness to slang, it is not without error. It occasionally fails on highly nuanced or figurative expressions, or in cases of overlapping or multi-token entities—highlighting the ongoing need for more domain-specific fine-tuning and interpretability tools in opioid-related NER. Overall, XLM-R offers significant improvements, but its limitations remind us that deep learning models in clinical NLP must be evaluated critically, especially when applied to noisy and emotionally charged social media text.

## V. Discussion

The comparative analysis of machine learning, deep learning, and transformer-based models on the Opioid Named Entity Recognition (ONER) task reveals several critical insights with substantial practical implications for both research and real-world applications.

### A. Practical Implications

Firstly, the superior performance of transformer-based models such as BERT-base-NER and RoBERTa-base—with precision, recall, F1-score, and accuracy all near 0.97—demonstrates the transformative impact that contextualized language representations have on domain-specific NLP tasks. This improvement is particularly meaningful in the context of opioid-related social media data, which is often noisy, informal, and linguistically diverse. The ability of these models to effectively disambiguate slang, fragmented sentences, and polysemous terms ensures more reliable detection of opioid-related entities, which can significantly enhance automated monitoring and intervention systems.

Secondly, the findings highlight the limitations of traditional machine learning and even some deep learning models when confronted with the linguistic complexity inherent to social media discourse. Random Forest and BiLSTM models, while capable, lag behind transformers in balancing precision and recall. This suggests that applications relying solely on classical approaches may risk higher rates of false positives or negatives, reducing their efficacy in sensitive domains such as drug abuse monitoring.

### B. Applications

The ONER-2025 dataset and the accompanying high-performing transformer models pave the way for numerous impactful applications:

- Public Health Surveillance: Automated, accurate extraction of opioid-related information from platforms like Reddit can support real-time monitoring of drug trends, emerging substances, and risk behaviors. Health authorities and policymakers could use these insights to inform timely public health interventions and resource allocation.
- Clinical Decision Support: Integrating opioid NER models into electronic health record systems could help clinicians better identify patients' drug histories and symptoms related to opioid use, enabling more personalized and safer treatment plans.
- Online Moderation and Harm Reduction: Social media platforms and harm reduction organizations could deploy these models to detect and flag risky or distressing content, facilitate outreach, and promote recovery-oriented messaging within online communities.
- Research and Epidemiology: The annotated ONER-2025 dataset itself serves as a valuable resource for further studies in opioid abuse patterns, language use, and intervention strategies, especially given its detailed entity categories and careful annotation.

### C. Importance of the Findings and Limitations

The significance of this work extends beyond methodological advancement. It addresses a pressing societal issue by equipping stakeholders with tools to analyze and interpret opioid-related discourse at scale. By successfully adapting state-of-the-art transformer architectures to this challenging domain, the study bridges the gap between technical NLP research and urgent public health needs.

Moreover, the hybrid annotation methodology, combining manual expertise with model-assisted labeling, showcases an efficient pathway to generate large-scale, high-quality domain-specific datasets—a critical step toward improving model generalizability and trustworthiness in medical NLP applications.

Finally, the statistical validation through paired t-tests affirms the robustness of transformer-based improvements over baseline models, reinforcing confidence in deploying such models in practical settings. This evidence-based approach underscores the reliability of the proposed models for critical real-world tasks.

However, several limitations are worth noting. While transformer models achieved high accuracy, their computational cost and inference latency pose major challenges for real-time implementation. The proposed real-time monitoring system, though promising for public health surveillance, requires substantial infrastructure to deploy models like BioBERT or ClinicalBERT on streaming data. Additionally, integrating data from heterogeneous sources—such as social media, electronic health records, and emergency response systems—raises non-trivial challenges in data harmonization, standardization, and scalability. In addition, ethical considerations such as user privacy, informed consent, and the responsible use of social media data must be carefully addressed when designing such monitoring systems.

## VI. Conclusion and Future work

In conclusion, this study highlights the critical role of Natural Language Processing (NLP), specifically Opioid Named Entity Recognition (ONER), in analyzing unstructured data from social media platforms such as Reddit. By developing the ONER-2025 dataset and addressing key challenges such as drug-related slang, ambiguous terms, fragmented sentences, and emotionally charged language , within opioid overdose discussions, we have provided a valuable resource for understanding opioid misuse discussions. Our findings demonstrate that transformer-based models (bert-base-NER, roberta-based) are effective in handling non-standard opioid-related terminology, achieving 97% accuracy and F1-score. This performance underscores the ability of advanced NLP techniques to correctly interpret complex user discourse, even in the absence of direct mentions of opioid names. Furthermore, by identifying patterns in opioid-related discussions, this research contributes to real-time monitoring of opioid misuse and provides healthcare professionals and policymakers with critical insights. Acknowledging these linguistic challenges and refining NLP approaches will enable better detection of opioid-related risks and improve public health interventions. Ultimately, this study advances social media-based opioid surveillance by demonstrating how NER models can adapt to evolving language patterns, ensuring more accurate and meaningful analysis of opioid discussions online.

Future research could expand the ONER-2025 dataset to include data from other social media platforms and improve the annotation process with domain experts. We also aim to develop hybrid models that combine rule-based and machine learning techniques to enhance accuracy. Additionally, integrating multi-modal data and applying real-time NER models could help monitor opioid-related trends and support targeted public health interventions. By advancing ONER-2025 technologies, we aim to assist in combating the opioid crisis and improving public health outcomes.

## Acknowledgment

The work was done with partial support from the Mexican Government through the grant A1-S-47854 of CONAHCYT, Mexico, grants 20241816, 20241819, and 20240951 of the Secretaría de Investigación y Posgrado of the InstitutoPolitécnicoNacional, Mexico. The authors thank the CONAHCYT for the computing resources brought to them through the Plataforma de AprendizajeProfundo para Tecnologías del Lenguaje of the Laboratorio de Supercómputo of the INAOE, Mexico and acknowledge the support of Microsoft through the Microsoft Latin America PhD Award.

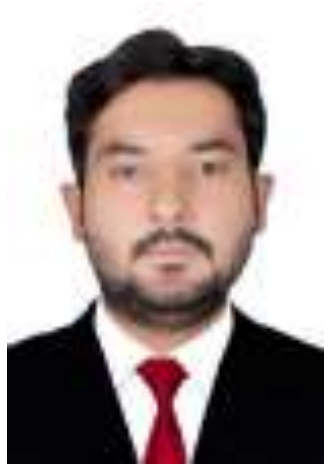

**Muhammad Ahmad** (Member, IEEE) received the B.S. degree in computer science from COMSATS University, in 2016, and the master's degree in computer science from Lahore Leads University, Lahore, Pakistan, in 2020. He is currently pursuing the Ph.D. degree in computer science with Centro de Investigación en Computación, Instituto Politécnico Nacional (CIC-IPN), Mexico City, with a specialization in computational linguistics, and automatic natural language processing. He is a Lecturer with The Islamia University of Bahawalpur, Pakistan. His research interests include health data science, opioid crisis detection, deep learning, large language models, and application of machine learning methods to natural language processing tasks.

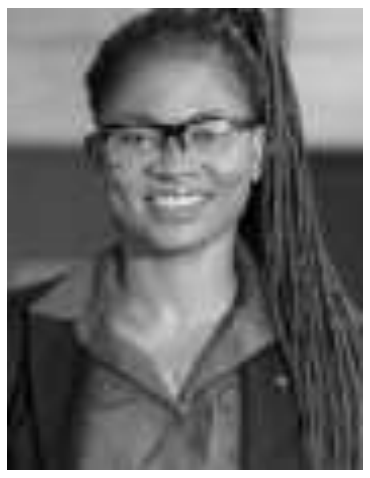

**Rita Orji**. (Member, IEEE) is currently a Computer Science Professor with Dalhousie University, Canada. Her research areas include human–computer interaction, persuasive technology, games for change, and digital health. She is well-known for her work in the area of personalizing persuasive technology and has published over 100 peer-reviewed articles in this area. Her research group is particularly interested in investigating user-centered approaches to designing interactive systems to motivate people for actions and causes that are beneficial for them and their communities as well as how interactive systems can be designed for the under-served population (HCI for Development (HCI4D)).

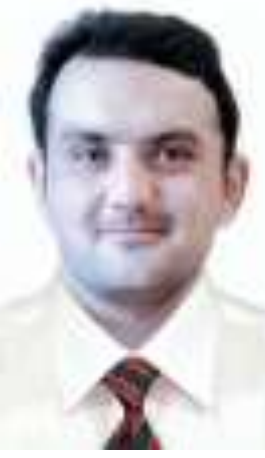

**Fida Ullah.** Received the master's degree in computer science from Beijing University of Chemical Technology, China, in 2021. He is currently pursuing the Ph.D. degree in computer science with Centro de Investigación en Computación, Instituto Politécnico Nacional (CICIPN), Mexico City, with a specialization in computational linguistics and automatic natural language processing. His research interests include named entity recognition, deep learning, large language models, and natural language processing.

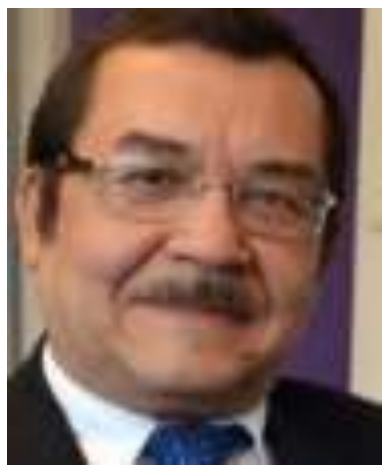

**Ildar Batyrhsin.** (Senior Member, IEEE) received the degree from the Moscow PhysicalTechnical Institute (MIPT), Faculty of Control and Applied Mathematics, the Ph.D. degree from the Moscow Power Engineering Institute (MPEI TU), and the Dr.Sc. (Habilitation) degree from the Higher Attestation Committee of Russian Federation. After finishing MIPT, he joined the Department of Informatics and Applied Mathematics, Kazan State Technological University (currently National Research Technological University), Kazan, Russia, and the Department Head, from 1997 to 2003. Since 1999, he has been also a Leading Researcher with the Institute of Problems of Informatics, Academy of Sciences of the Republic of Tatarstan, Russia. In 2003, he joined the Research Program of Applied Mathematics and Computations of Mexican Petroleum Institute (IMP) as a Distinguished Invited Researcher and later as a Leading Researcher, the Head of project, and a Professor of postgraduate education. Since 2014, he has been the Titular Professor C of Centro de Investigación en Computación, Instituto Politécnico Nacional (CIC-IPN), Mexico City. He is a member of the IEEE SMC Board of Governors, a Senior Member of IEEE CI and SMC Societies, a member of the Management Committee of IEEE TRANSACTIONS ON ARTIFICIAL INTELLIGENCE, and a member of the Award Committee of IEEE Lotfi A. Zadeh Award for Emerging Technologies. He is the PC Co-Chair of the IEEE SMC 2023 Conference in Hawaii, USA.

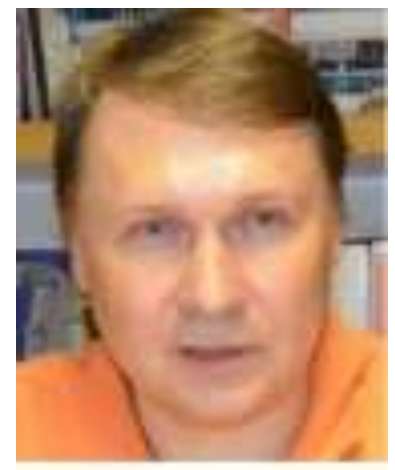

**Grigori Sidorov.** Received the Doctor of Science (Ph.D.) degree from the ‘‘Lomonosov’’ State University of Moscow, Russia, in 1996. He is currently a Professor and a Researcher with the Natural Language and Text Processing Laboratory, Centro de Investigación en Computación, Instituto Politécnico Nacional (CIC-IPN), Mexico City. He is the author of more than 180 publications in computational linguistics and automatic natural language processing. His works have been cited more than 400 times (without self-citations). His scientific research interests include computational linguistics, automatic text processing, and the application of machine learning methods to natural language processing tasks. He is a National Researcher of Mexico (a member of the SNI) level 3, a member of the Mexican Academy of Sciences, and the Editor-in-Chief of the research journal Computación y Sistemas (included in ISI and Scopus, among other indices).